\documentclass[letterpaper, 10 pt, conference]{ieeeconf}  
                                                           
\IEEEoverridecommandlockouts

\usepackage{amsmath} 
\usepackage{amssymb}  

\usepackage{graphicx, xcolor, wasysym}

\title{\LARGE \bf
Dynamic Coupling and Indirect Control of Jointed Robots Rolling Atop A Moving Platform
}

\author{Hamidreza Moradi and Scott David Kelly
\thanks{This material is based upon work supported by the U.S. National Science Foundation under Grant No. 2140118.}
\thanks{Both authors are in the Department of Mechanical Engineering and Engineering Science, 
 	University of North Carolina at Charlotte,
	Charlotte, NC 28223 USA {\tt\small hmoradi@charlotte.edu}, {\tt\small scott.kelly@charlotte.edu}}
}

\begin{document}

\maketitle
\thispagestyle{empty}
\pagestyle{empty}


\begin{abstract}

An asymmetric two-link robot supported atop a flat platform by wheels that roll and pivot freely, but do not slip laterally, will develop forward momentum if the joint between the links is actuated internally. In particular, oscillations in the joint angle will generate undulatory locomotion suggesting fishlike swimming. If two such robots surmount a common platform that's free to translate with its own inertial dynamics, then the individual robots' dynamics will be coupled so that the locomotion of either robot is affected by that of the other. We develop a mathematical model for this system and present simulations demonstrating its behavior. We then consider a single robot with an unactuated joint rolling atop a platform that moves under control, and show that actuation of the platform is sufficient to dictate the robot's behavior. In particular, with the acceleration of the platform as an input, the robot's heading can be made to track a chosen function of time. This is sufficient to guarantee that the robot can be induced to orbit a fixed point on the platform or to locomote persistently in a desired direction.

\end{abstract}


\section{Introduction}

\begin{figure}
\centering
\includegraphics[width = 0.25\textwidth]{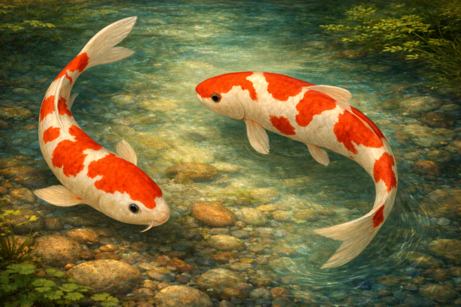} \\
\includegraphics[width = 0.4\textwidth]{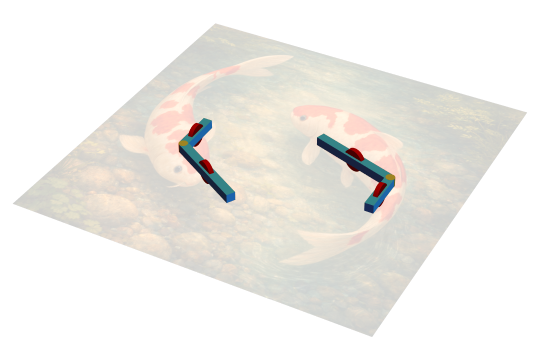}
\caption{Pairs of agents that flex their bodies for propulsion and steering, coupled through the inertial dynamics of shared environmental media.} 
\label{twoFish.fig}
\end{figure}

Fig. \ref{twoFish.fig} suggests an analogy between two mechanical systems, one a very simplified model for the other. Like each of the fish in the top panel, each of the wheeled robots in the bottom panel can flex its body at a point aft of its midline. For fish or for robot, flexing in this way generates forward propulsion, producing a force on the environment in the opposite direction. After a period of undulatory motion to build forward momentum, either fish or robot can cease to undulate yet be carried along thereafter by this momentum, flexing only to steer. If the platform in the bottom panel is free to translate with finite mass of its own, then both systems shown have the property that the movement of either individual will excite the environment dynamically, affecting the dynamics of the other individual. Either individual can propel itself and navigate in such a way as to help or to hinder the propulsion or navigation of the other.

It's well known that schooling fish can swim mutualistically, enjoying greater energy efficiency en masse than they'd exhibit if their individual dynamics were decoupled \cite{zhanglauder24}, but the dynamics interior to fish schools are subtle. Experiments with a self-propelling array of flapping paddles, driven individually to execute the same periodic flapping motion, have shown that propulsive efficiency --- or, differently, propulsive speed --- can be improved by selecting particular patterns of spacing and phasing among the paddles \cite{bhansali18}. This suggests that mutualistic efficiency could be an emergent property of fish schools in which individuals propel themselves actively without particular regard for the hydrodynamics of their neighbors. On the other hand, it's been shown experimentally that the flexible body of a deceased fish can be stimulated to execute a forward swimming motion, against the steady background flow in a water channel, by the vortices shed from a bluff body upstream \cite{bhtll06}. This suggests that hydrodynamic drafting could play a role in efficient schooling, whereby an individual fish might experience a relaxed responsibility for dictating its own swimming motion by allowing its body to be driven by the ambient flow.

In the present paper, we explore both of these ideas using the system in the lower panel of Fig. \ref{twoFish.fig}. 
In Section \ref{amsModel.sec} we present a model for a solitary two-link robot rolling freely atop a stationary platform with direct control over its internal joint, and we demonstrate that joint actuation generically produces forward locomotion.
In Section \ref{amsSchooling.sec} we proceed to a model for two such robots sharing a common platform with inertial dynamics of its own, and show that the relative phasing of the robots' oscillations affect their collective propulsion.
In Section \ref{passiveRobot.sec} we relax the assumption of direct joint actuation and model the dynamics of a robot with a passive joint atop a driven platform. We show that with the acceleration of the platform as a control input, the robot's heading can be induced to track a desired function of time. We suppose the robot to have some elasticity at its joint, pushing it toward a straightened state, to ensure that the ``fully folded'' configuration isn't an attractive internal equilibrium. In this case, the robot can be made to navigate the platform arbitrarily via heading control.

Note that the authors don't claim the dynamics of the two systems in Fig. \ref{twoFish.fig} to be similar quantitatively. Neither is it true, however, that the upper system is unique in manifesting the idea of a practical multi-agent scenario in which agents are coupled through the responsive physics of a shared medium. Some principles of analysis or control may be appropriate for all such systems; certainly others will not. The present paper investigates the topic in the context of nonholonomic mechanics, exploiting a combination of different modeling paradigms and tools from differential geometry to expose the dynamics and enable the control of a biologically inspired system.


\section{Dynamics of a solitary robot on a stationary platform with direct joint control}
\label{amsModel.sec}

\begin{figure}
\centering
\includegraphics[width = 0.25\textwidth]{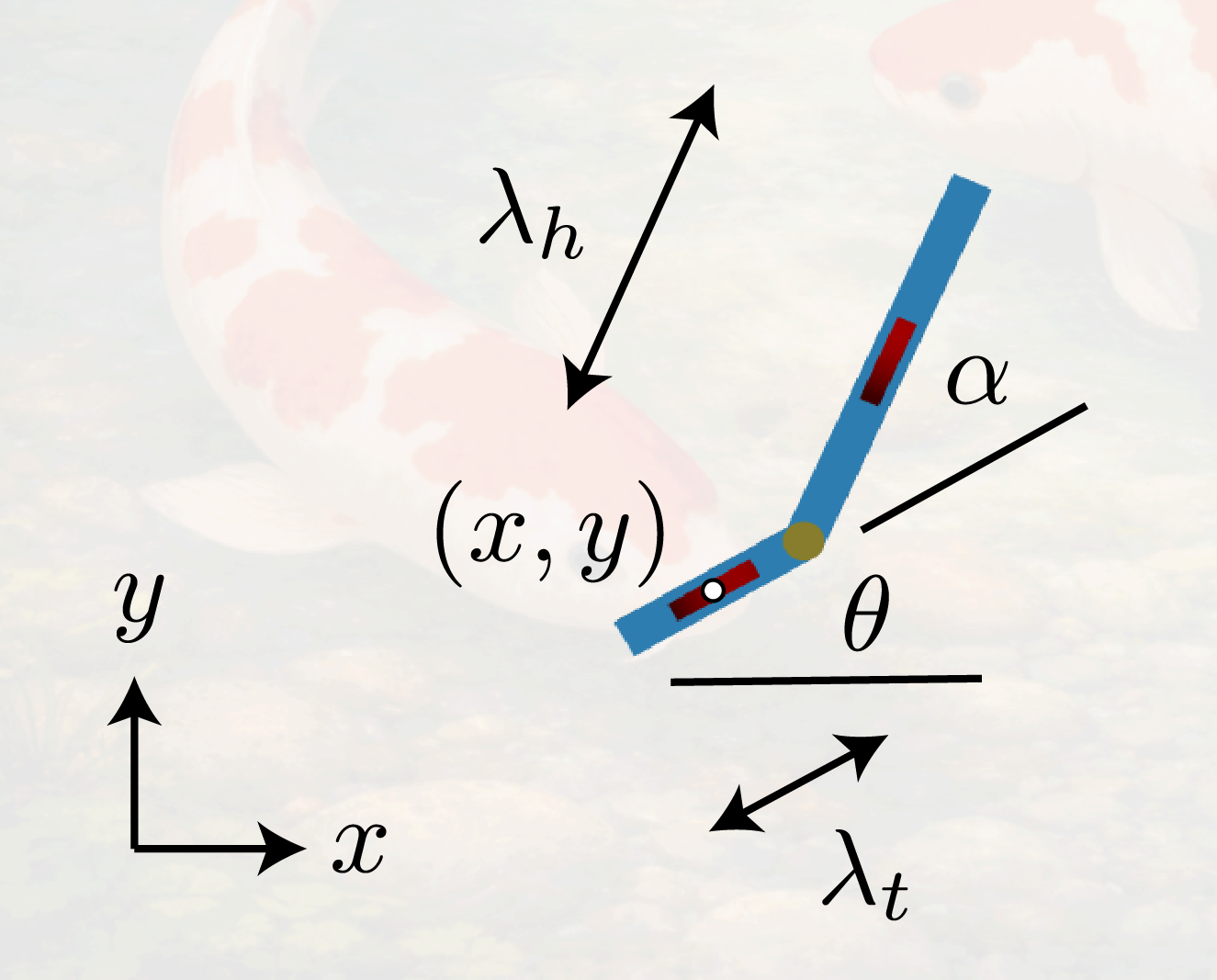}
\caption{Parameterization of a single robot.}
\label{oneFish.fig}
\end{figure}

Each of the robots in the lower panel of Fig. \ref{twoFish.fig} consists of two rigid links supported by wheels at their midpoints. The wheels are free to roll longitudinally and to pivot but not to slip laterally. Fig. \ref{oneFish.fig} defines coordinates $(\alpha, x, y, \theta)$ on a single robot's configuration manifold $Q$. We'll denote the mass, length, and rotational inertia of each link by $m$, $\lambda$, and $J$, respectively, with subscripts $h$ and $t$ for ``head'' (the longer link) and ``tail''. Prohibiting the wheels from slipping laterally is equivalent to requiring that the system's velocity in $Q$ be annihilated at all times by the one forms
\begin{align*}
\omega_t &= - (\sin \theta) \, dx + (\cos \theta) \,dy, \\
\omega_h &=
\frac{\lambda}{2} \, d\alpha - \sin (\theta + \alpha) \, dx \\ 
& \qquad \qquad + \cos (\theta + \alpha) \, dy + \frac{\lambda}{2} (\lambda_h + \lambda_t \cos \alpha) \, d \theta.
\end{align*}
These are nonholonomic constraints, and with the system's kinetic energy as a Lagrangian, the dynamics in $(\alpha, x, y, \theta)$ are nominally given by Euler-Lagrange equations with Lagrange multipliers representing the constraint forces. The dynamics can be expressed more compactly, however, by exploiting the system's symmetry with respect to planar translation and reorientation --- that is, by noting that the Lagrangian and the constraints, and therefore the dynamics, are expressible in terms of velocities in a body-coincident frame of reference. 

Instantaneously, the robot's motion is always a superposition of flows along the two vector fields
\begin{align*}
\circlearrowleft \, =&  \left(\lambda_h + \lambda_t \cos \alpha \right) \left(\frac{\cos \theta}{2 \sin \alpha}\right) \frac{\partial}{\partial x} \\
& + \left(\lambda_h + \lambda_t \cos \alpha \right) \left(\frac{\sin \theta}{2 \sin \alpha}\right) \frac{\partial}{\partial y} +  \frac{\partial}{\partial \theta}, \\
\curlyvee =& \frac{\partial}{\partial \alpha} + \left(\frac{4 \lambda_h \sin \alpha \cos \theta}{\Delta}\right) \frac{\partial}{\partial x} + \left(\frac{4 \lambda_h \sin \alpha \sin \theta}{\Delta}\right) \frac{\partial}{\partial y} \\
&- \left(\frac{2 \lambda_h (\lambda_h + \lambda_t \cos \alpha)}{\Delta}\right) \frac{\partial}{\partial \theta}
\end{align*}
on $Q$, where
\begin{equation*}
\Delta = 4 + 2 \lambda_h^2 + \lambda_t^2 + 4 \lambda_h \lambda_t \cos \alpha + \left(\lambda_t^2 - 4\right) \cos (2 \alpha).
\end{equation*}
Together these vector fields span the distribution on $Q$ that's annihilated by $\omega_t$ and $\omega_h$. Flow along $\circlearrowleft$ corresponds to rolling longitudinally along a circular arc with $\alpha$ held fixed. Flow along $\curlyvee$ corresponds to the scissoring motion achieved when $\alpha$ is varied with the robot initially at rest.
Following \cite{bkmm96}, a scalar \emph{nonholonomic momentum} $p$ can be associated with flow along $\circlearrowleft$ by defining
\begin{equation*}
p = \left\langle \frac{\partial L}{\partial \dot q}, \circlearrowleft \right\rangle,
\end{equation*}
where $\langle \cdot, \cdot \rangle$ denotes the natural pairing of cotangent and tangent vectors on $Q$. The quantity $p$ represents angular momentum with respect to the center of rotation defined by $\circlearrowleft$. The constrained dynamics in $(\alpha, x, y, \theta)$ reduce to unconstrained dynamics in $(\alpha, p)$ in the sense that if $\alpha$ and $p$ are known as functions of time, then $x$, $y$, $\theta$ can be reconstructed as functions of time from their initial conditions. Explicit calculation shows that
\begin{equation*}
p = J_h \dot \alpha + \frac{1}{4} \left(\varrho_x \dot x + \varrho_y \dot y + \varrho_\theta \dot \theta\right),
\end{equation*}
where
\begin{align*}
\varrho_x &= \frac{1}{\sin \alpha} \big( \left(2 (m_h + m_t) \lambda_t \cos \alpha\right. \\
&\qquad \qquad \quad + \left.\lambda_h (m_h + 2 m_t + m_h \cos 2 \alpha)\right) \cos \theta \\
 &\qquad \qquad \quad - 2 m_h (\lambda_t + \lambda_h \cos \alpha) \sin \alpha \sin \theta \big), \\
 \varrho_y &= 2 m_h (\lambda_t + \lambda_h \cos \alpha) \cos \theta + (2 (m_h + m_t) \lambda_t \cos \alpha \\
 &\qquad \qquad \quad + \lambda_h (m_h + 2 m_t + m_h \cos 2 \alpha))
\, \frac{\sin \theta}{\sin \alpha}, \\
\varrho_\theta &= 4 (J_h + J_t) + m_h \lambda_t^2 + m_h \lambda_h \lambda_t \cos \alpha.
\end{align*}

The ODE governing the dynamics of $p$ can be obtained by considering the system's symmetry with respect to planar translation and reorientation in a more technical sense. This symmetry corresponds to an invariance of the Lagrangian and the constraint forms under an action of the Lie group $SE(2)$ on the manifold $Q$. Any vector field on $Q$ that's everywhere tangent to orbits of the group action --- that is, any vector field on $Q$ that lacks a component in the $\alpha$ direction --- will coincide at each point $q \in Q$ with an \emph{infinitesimal generator} of the action. Infinitesimal generators are a family of vector fields $\xi_Q$ on $Q$ in one-to-one correspondence with Lie algebra elements $\xi$. The vector field $\circlearrowleft$ thus corresponds at each point $q = (\alpha, x, y, \theta)$ to an infinitesimal generator $\xi_Q$ associated with some $q$-dependent $\xi \in \mathfrak{se}(2)$. A formula for $\xi(q)$ can be computed directly (in a manner detailed in \cite{bkmm96}) from the formula for $\circlearrowleft (q)$. The dynamics of $p$ then satisfy
\begin{equation}
\dot p
=
\left\langle \frac{\partial L}{\partial \dot q},
\left[\frac{d\xi (q)}{dt}\right]_Q\right\rangle.
\label{eq:dotP}
\end{equation}
Performing these calculations explicitly, we obtain
\begin{equation}
\dot p = \frac{\sin \alpha}{4 \earth} \big(\mars(\alpha) p + \venus(\alpha) \dot \alpha\big) \dot \alpha,
\label{pdot.eq}
\end{equation}
where 
\begin{align*}
\mars(\alpha) &= \\
&\mkern-30mu - 4 (m_h + m_t) \left(2 \left(\lambda_h^2 + \lambda_t^2\right) \cos \alpha + \lambda_h \lambda_t \left(3 + \cos 2 \alpha\right)\right), \\
\venus(\alpha) &= - 8 (m_h + m_t) \left(J_t \lambda_h^2 - J_h \lambda_t^2 \right) \cos \alpha \\
&+ \lambda_h \lambda_t \big(4 J_h (2 m_h + m_t) - 4 J_t (m_h + 2 m_t) \\
&+ m_h m_t (\lambda_h - \lambda_t) (\lambda_h + \lambda_t) \\
&+ \left(4 J_h m_t - 4 J_t m_h + m_h m_t \left(\lambda_t^2 - \lambda_h^2\right)\right) \cos 2 \alpha\big), \\
\earth = \, & 4 J_h + 4 J_t + (m_h + 2 m_t) \lambda_h^2 + (2 m_h + m_t) \lambda_t^2 \\
&+ 4 (m_h + m_t) \lambda_h \lambda_t \cos \alpha \\
&+ \left(m_h \lambda_h^2 + m_t \lambda_t^2 - 4 J_h - 4 J_t \right) \cos 2 \alpha.
\end{align*}

It's apparent from \eqref{pdot.eq} that as long as $\venus(\alpha) \ne 0$, actuation of the robot's joint will induce the robot to roll longitudinally --- in particular, forward --- even from rest. This condition is satisfied as long as it isn't simultaneously the case that $m_t = m_h$, $\lambda_t = \lambda_h$, and $J_t = J_h$. Those three equalities together would imply front-to-back symmetry of the robot, so that variations in $\alpha$ would cause the robot initially at rest to scissor along $\curlyvee$ --- its joint moving along the straight bisector of $\alpha$ --- without beginning to roll lengthwise.

If $\dot \alpha$ is specified as a control input, then \eqref{pdot.eq} models the robot's dynamics completely. Its dynamics in $(x, y, \theta)$ can be reconstructed from its dynamics in $(\alpha, p)$ via the equations 
\begin{align*}
\dot x &= \frac{\cos \theta \sin \alpha}{\earth}  \left(4 \left(\lambda_h + \lambda_t \cos \alpha\right) p + 
\left(\lambda_h \left(4 J_t + m_h \lambda_t^2\right) \right.\right.\\
& \qquad \qquad \qquad +\left.\left.\left(m_h \lambda_h^2 - 4 J_h\right) \lambda_t \cos\right) \dot \alpha \right), \\
\dot y &= \frac{\sin \theta \sin \alpha}{\earth} \left(4 \left(\lambda_h + \lambda_t \cos \alpha\right) p + 
\left(\lambda_h \left(4 J_t + m_h \lambda_t^2\right) \right.\right. \\
& \qquad \qquad \qquad +\left.\left.\left(m_h \lambda_h^2 - 4 J_h\right) \lambda_t \cos\right) \dot \alpha \right), \\
\dot \theta &= \frac{1}{\earth} \bigg(8 p \left(\sin \alpha\right)^2 - \big(4 J_h + (m_h + 2 m_t) \lambda_h^2 \\
&+ 2 (m_h + m_t) \lambda_h \lambda_t \cos \alpha + \left(m_h \lambda_h^2 - 4 J_h\right) \cos 2 \alpha \big)\dot \alpha \bigg).
\end{align*}


\section{Coupled dynamics of robots with direct joint control sharing a movable platform}
\label{amsSchooling.sec}

Suppose that the platform beneath the robot from Section \ref{amsModel.sec} had some externally dictated translational velocity. The equations of motion for the robot would be altered by the incorporation of this additional velocity into the robot's Lagrangian. If the variables $(x, y, \theta)$ are understood to define the robot's position and orientation with respect to the platform, then the constraint forms $\omega_t$ and $\omega_h$ would be unchanged. If the platform's velocity were expressed relative to a body-coincident frame --- say, $f(t)$ along the tail and $g(t)$ perpendicular to the tail --- then the Lagrangian would obviously retain the symmetry present in the fixed-platform case, and the process of symmetry reduction would yield a reduced model for the robot's dynamics as in Section \ref{amsModel.sec}.

A model for multiple robots interacting through the inertial dynamics of a common platform that's free to translate can be constructed in a modular way by first obtaining the reduced equations of motion and the reconstruction equations for a single robot on a moving platform, in the manner described above, and then replicating this set of equations for each robot in the system. Assigned to each robot $i$ is a distinct heading $\theta_i(t)$ and a distinct environmental velocity with components $(f_i(t), g_i(t))$. Now suppose that the platform has finite mass $M$ and displacement $(X, Y)$ with respect to a stationary external reference frame. For each robot, we assign
\begin{equation*}
f_i(t) = \dot X \cos \theta_i + \dot Y \sin \theta_i, \quad
g_i(t) = - \dot X \sin \theta_i + \dot Y \cos \theta_i.
\end{equation*}
The total translational momentum in the system along the $X$ and $Y$ directions --- relative to the stationary frame --- is just the sum of the individual momenta of the robots' links together with the momentum of the platform. If no external force acts on the system, then this total momentum will remain constant over time. If the platform and all robots are initially at rest, then the total momentum will be zero for all time. Coupled with the sets of ODEs governing the robots' motion relative to the platform, the equation stipulating conservation of momentum is sufficient to model the dynamics of the system overall, driven by the angular velocities at the robots' joints as inputs.

\begin{figure}
\centering
\includegraphics[width = 0.5\textwidth]{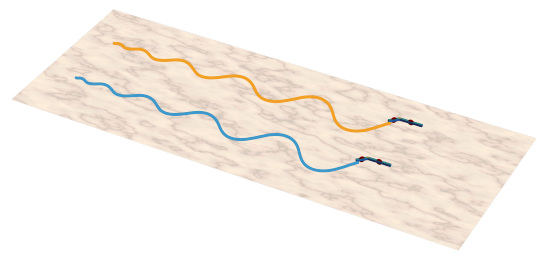}
\caption{Synchronized forward locomotion from rest by two robots on a shared movable platform, with $M = 1$, $m_t = \lambda_t = 1$, $m_h = \lambda_h = 2$, and $J = m \lambda^2 / 4$ for each link.}
\label{schoolOfTwo.fig}
\end{figure}

Oscillations of $\alpha_i$ around zero will induce the $i$th robot to move forward in an undulatory way relative to the platform. Fig. \ref{schoolOfTwo.fig} depicts the trajectories of two robots accelerating side by side from rest with $\alpha_1 = \alpha_2 = \frac{\pi}{4} \cos t$. The plate is initially at rest as well, so that the net translational momentum in the system remains zero throughout the simulation. As each robot advances, it pushes the medium of the platform backwards, just as a swimming fish imparts backward momentum to the water in its wake.

\begin{figure}
\centering
\includegraphics[width = 0.42\textwidth]{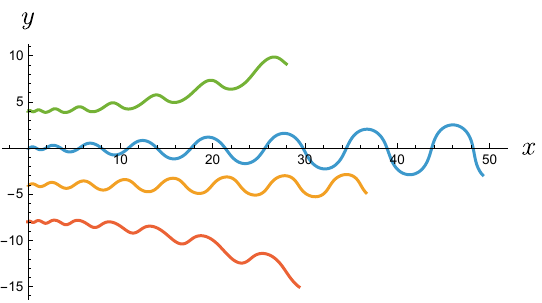}
\caption{Trajectories relative to the platform for one of the robots in Fig. \ref{schoolOfTwo.fig}, in every case with $\alpha = \frac{\pi}{4} \cos t$ over the same period of time, corresponding to variations in the heading or chirality of the other robot. The trajectory generating the greatest final displacement corresponds to synchronized parallel locomotion as in Fig. \ref{schoolOfTwo.fig}. The adjacent straight trajectory corresponds to the situation in which the second robot mirrors the first, with $\alpha = -\frac{\pi}{4} \cos t$. The curved trajectories correspond to the second robot's propelling itself in different directions.}
\label{helpOrHinder.fig}
\end{figure}

The motion of either robot in Fig. \ref{schoolOfTwo.fig} relative to the platform will influence the motion of the other. Fig. \ref{helpOrHinder.fig} illustrates this point by depicting four possible trajectories that one of the robots might describe while executing eight cycles of the sinusoidal oscillation in shape from Fig. \ref{schoolOfTwo.fig}. Differences result from changes in what the other robot is doing: its average heading is different, its gait is time-shifted to be half a cycle out of phase, or both.


\section{Heading control for a passive robot on an actuated platform}
\label{passiveRobot.sec}

We return now to the case in which only a single robot is present, but we suppose the robot's joint to be unactuated, its dynamics driven by our prescribing the acceleration of the platform. If $\alpha(t)$ isn't specified as a control input, then a second-order ODE is required to govern its evolution. The correct ODE can be derived in a variety of ways. The formalism from \cite{bkmm96} that we used in Section \ref{amsModel.sec} accommodates mechanical systems with free internal dynamics, and it's possible to follow that formalism to model the dynamics of $\alpha$ in terms of the nonholonomic momentum $p$. On the other hand, in a previous paper in which the authors analyzed the free dynamics of a back-to-front symmetric robot on a stationary platform \cite{kellymoradi25}, we found it more straightforward to use the Hamiltonian reduction method developed in \cite{batessniatycki93} to realize a set of equations modeling the system overall. The results appearing in the remainder of the present paper are based on a model obtained by combining Gibbs-Appell formalism \cite{qztso22} with symmetry reduction in a novel way, summarized below. This approach exposes the system to the straightforward addition of control inputs in a manner that the Hamiltonian approach of \cite{kellymoradi25} would not.

\begin{figure}
\begin{minipage}{0.26 \textwidth}
\includegraphics[width = \textwidth]{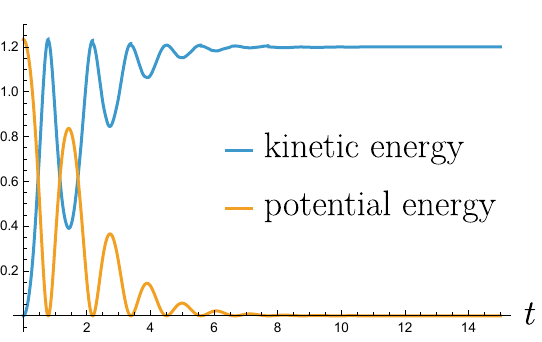}
\end{minipage} \quad
\begin{minipage}{0.2 \textwidth}
\includegraphics[width = \textwidth]{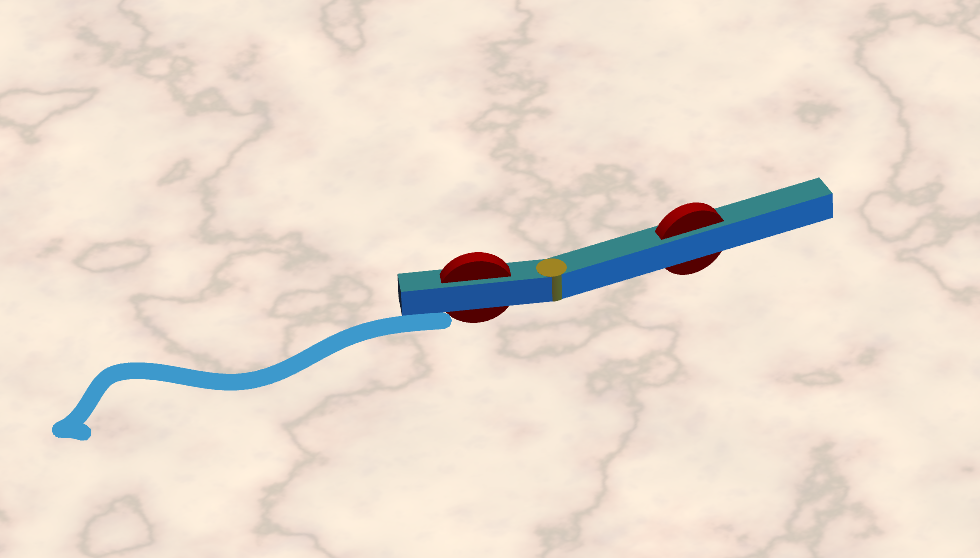}
\end{minipage}
\caption{With a torsional spring at its joint to pull the joint toward $\alpha = 0$, the robot set free on a stationary platform with $\alpha \ne 0$ will convert all energy initially stored in the spring into translational kinetic energy. There's no damping at the joint; oscillations in $\alpha$ decay in amplitude over time as a result of nonholonomic mechanics.}
\label{unsprung.fig}
\end{figure}

We make one additional assumption before proceeding. We remove actuation from the robot's joint, but we suppose some elasticity to be present at the joint --- modeled as a linear torsional spring --- with the tendency to restore the joint to its straightened configuration. This serves to guarantee that if our indirect control steers the robot to a completely folded state with $\alpha = \pm \pi$, this state won't persist. As was remarked of the actuated robot, the spring-loaded robot exhibits a bias for forward locomotion. Released from rest with $\alpha \ne 0$ on a stationary platform, the robot will convert all energy initially stored in the spring into translational kinetic energy. Undulations in the joint will diminish in amplitude toward zero over time as the robot approaches a state of rolling in a straight line. This is illustrated in Fig. \ref{unsprung.fig}.


Regardless of the placement of actuation, the mechanics of a solitary robot and a movable platform, taken together as a single system, retain the SE(2) symmetry described in Section \ref{amsModel.sec}, since we can always define the platform velocity in terms of components relative to a body-coincident frame. We can therefore model the combined system with a concatenation of the reduced equations from Section \ref{amsModel.sec} for the robot's motion --- including the accommodation for moving contact points from Section \ref{amsSchooling.sec} --- together with a model for the platform's dynamics relative to a body-coincident frame.

This symmetry reduction is a natural
prerequisite for the Gibbs-Appell formulation. The Gibbs-Appell equations are written at the level of admissible accelerations/velocities and therefore require a representation of the system dynamics in terms of a minimal set of generalized variables that respect the nonholonomic constraints. The reduced momentum equation provides the dynamical evolution along the symmetry-generated direction, while the reconstruction equations lift the reduced trajectory back to the full tangent bundle. In effect, the reduction step identifies the appropriate generalized and pseudo-velocity structure of the moving-platform system, and the subsequent Gibbs-Appell formulation converts this geometric structure into a dynamical model suitable for controller design. This construction allows the platform accelerations to be treated as control inputs that influence the evolution of the passive joint and, consequently, the locomotion of the robot relative to the platform.

The Gibbs-Appell formulation provides a convenient framework because it operates directly at the level of admissible accelerations, avoiding the explicit appearance of forces associated with the nonholonomic constraints. With generalized coordinates $q$ and admissible accelerations $\ddot q$ that satisfy a set of nonholonomic velocity constraints, the Gibbs-Appell method introduces the \emph{Appellian} (or $S$-function)
\begin{equation*}
S(q,\dot q,\ddot q) = \frac{1}{2}\sum_{k} m_k \, a_k^2 + \frac{1}{2}\sum_{k} J_k \, \alpha_k^2,
\end{equation*}
where $a_k$ and $\alpha_k$ denote the linear and angular accelerations of the individual rigid bodies comprising the system. The equations of motion follow from the relations
\begin{equation}
\frac{\partial S}{\partial \ddot q_i} = Q_i ,
\label{eq:GAFunc}
\end{equation}
where $Q_i$ is the generalized force associated with the coordinate $q_i$. In this formulation the admissible accelerations automatically satisfy the nonholonomic constraints, and therefore the resulting equations describe the constrained dynamics without the need to introduce Lagrange multipliers.

In the present system the admissible accelerations are parameterized using the reduced description introduced earlier. Specifically, the reconstruction relations provide expressions for the velocities $(\dot x, \dot y, \dot \theta)$ in terms of the reduced variables $(\alpha, \dot \alpha, p)$ together with the platform velocities  $(\dot x_p, \dot y_p)$. The latter are defined relative to the body-coincident frame, so that they're the $(f, g)$ --- not the $(\dot X, \dot Y)$ --- of Section \ref{amsSchooling.sec}. 

Given the Appellian of the robot-platform system $S_{\mathrm{tot}}$, the evolution of the joint angle $\alpha$ follows from the Gibbs-Appell relation \eqref{eq:GAFunc}. Since we've equipped the joint with a linear torsional spring, the generalized force associated with $\alpha$ is $Q_\alpha = -k\,\alpha $.
Solving \eqref{eq:GAFunc} yields an explicit expression for the joint angular acceleration which, together with the expression for $\dot p$ obtained earlier, provides a closed description of the internal dynamics of the passive robot.

Our objective is heading control for the robot, but the heading variable $\theta$ isn't actuated directly. Instead, it evolves according to the reconstructed kinematics of the reduced model. Recall that the reduction procedure provides an explicit expression for the body angular velocity
\begin{equation}
\dot{\theta} = f_\theta(q_r, \dot{q}_r,\theta),
\label{eq:thetaReconstruction}
\end{equation}
where $q_r$ includes the reduced variables for the robot and platform both. Differentiating \eqref{eq:thetaReconstruction} and substituting for $\dot p$ from \eqref{eq:dotP} and $\ddot{\alpha}$ from \eqref{eq:GAFunc} produces an explicit acceleration-level equation for the heading in control-affine form
\begin{equation}
\ddot{\theta}
=
\Phi_\theta(q_r,\dot{q}_r,\theta,\dot\theta)
+
c_x(\cdot)\,\ddot{x}_p
+
c_y(\cdot)\,\ddot{y}_p ,
\label{eq:thetaAffineExpanded}
\end{equation}
where $\Phi_\theta(\cdot)$ contains all drift terms arising from the internal robot dynamics, and $c_x(\cdot)$ and $c_y(\cdot)$ are state-dependent coefficients multiplying the platform accelerations.

For the purpose of controller design it is convenient to rearrange \eqref{eq:thetaAffineExpanded} into the standard form
\begin{equation}
A_\theta(\cdot)\,\ddot{\theta}
+
F_{\theta,\mathrm{int}}(\cdot)
=
c_x(\cdot)\,\ddot{x}_p
+
c_y(\cdot)\,\ddot{y}_p
\label{eq:thetaStructured}
\end{equation}
with
\begin{align*}
A_\theta(\cdot) &= 
\frac{\partial}{\partial \ddot{\theta}}
\left(
\ddot{\theta}
-
\frac{d}{dt}\big(f_\theta(\cdot)\big)
\right), \\
F_{\theta,\mathrm{int}}(\cdot) &=
\ddot{\theta}
-
\Phi_\theta(\cdot)
-
A_\theta(\cdot)\ddot{\theta}.
\end{align*}
The structure shown in \eqref{eq:thetaStructured} allows us to use feedback linearization for heading control. Defining a desired heading trajectory $\theta_\text{desired}(t)$ and introducing the commanded angular acceleration
\begin{equation*}
\ddot{\theta}_{\mathrm{cmd}}
=
\ddot{\theta}_\text{desired}
-
d_1(\dot{\theta}-\dot{\theta}_\text{desired})
-
d_2(\theta-\theta_\text{desired}),
\label{eq:thetaCommand}
\end{equation*}
the auxiliary control variable
\begin{equation}
u_\theta
=
A_\theta(\cdot)\,\ddot{\theta}_{\mathrm{cmd}}
+
F_{\theta,\mathrm{int}}(\cdot)
\label{eq:thetaControlLaw}
\end{equation}
cancels the internal nonlinear dynamics. 

The inputs to the system don't act on $\theta$ directly in the sense of an unconstrained rigid body; rather, they act through the admissible dynamics of the robot and therefore through the internal joint dynamics encoded in $\alpha$. This means that, despite the apparent availability of two independent input channels, the heading control problem is effectively single-input at the level of the orientation dynamics. Consequently, one cannot in general assign two independent output objectives simultaneously. The platform accelerations must be coordinated to achieve a single dominant control goal, taken here to be the regulation or tracking of the heading angle $\theta$.

Among the infinitely many pairs $(\ddot{x}_p,\ddot{y}_p)$ satisfying \eqref{eq:thetaControlLaw}, a natural choice is the minimum-norm solution obtained by projecting the control action along the vector of input coefficients. This gives
\begin{equation}
\ddot{x}_p
=
\frac{u_\theta\, c_x(\cdot)}
{c_x(\cdot)^2 + c_y(\cdot)^2 + \varepsilon},
\qquad
\ddot{y}_p
=
\frac{u_\theta\, c_y(\cdot)}
{c_x(\cdot)^2 + c_y(\cdot)^2 + \varepsilon},
\label{eq:regularized-input}
\end{equation}
in which to avoid numerical difficulties when the control effectiveness becomes very small, we regularize the denominator using an extra $\varepsilon$ term. Substituting \eqref{eq:regularized-input} into \eqref{eq:thetaAffineExpanded} guarantees tracking of the desired heading angle.

\begin{figure}
    \centering
    \includegraphics[width=0.85\linewidth]{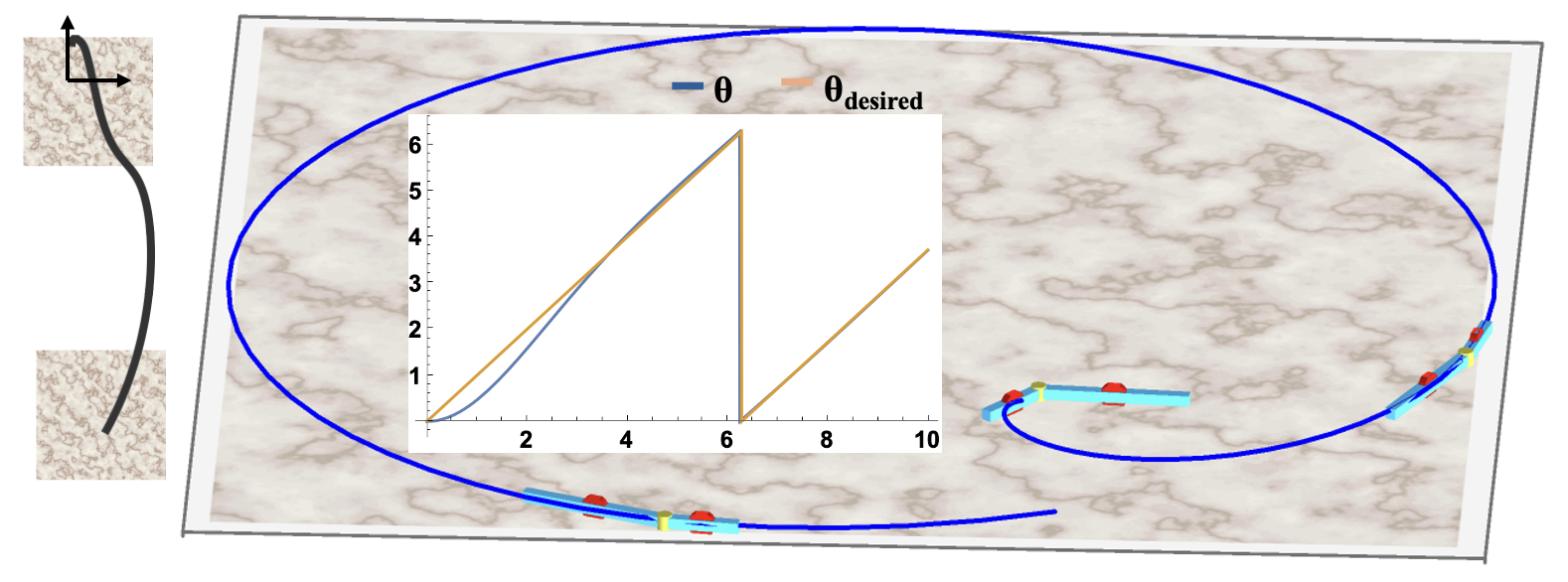}
    \caption{Stabilization of the robot to a circular trajectory, shown with three configuration snapshots. The inset plot depicts the tracking of $\theta_\text{desired}(t)=t$ (mod $2 \pi$). The platform motion is shown at left.}
    \label{fig:CirbleCont}
\end{figure}

\begin{figure}
    \centering
    \includegraphics[width=0.85\linewidth]{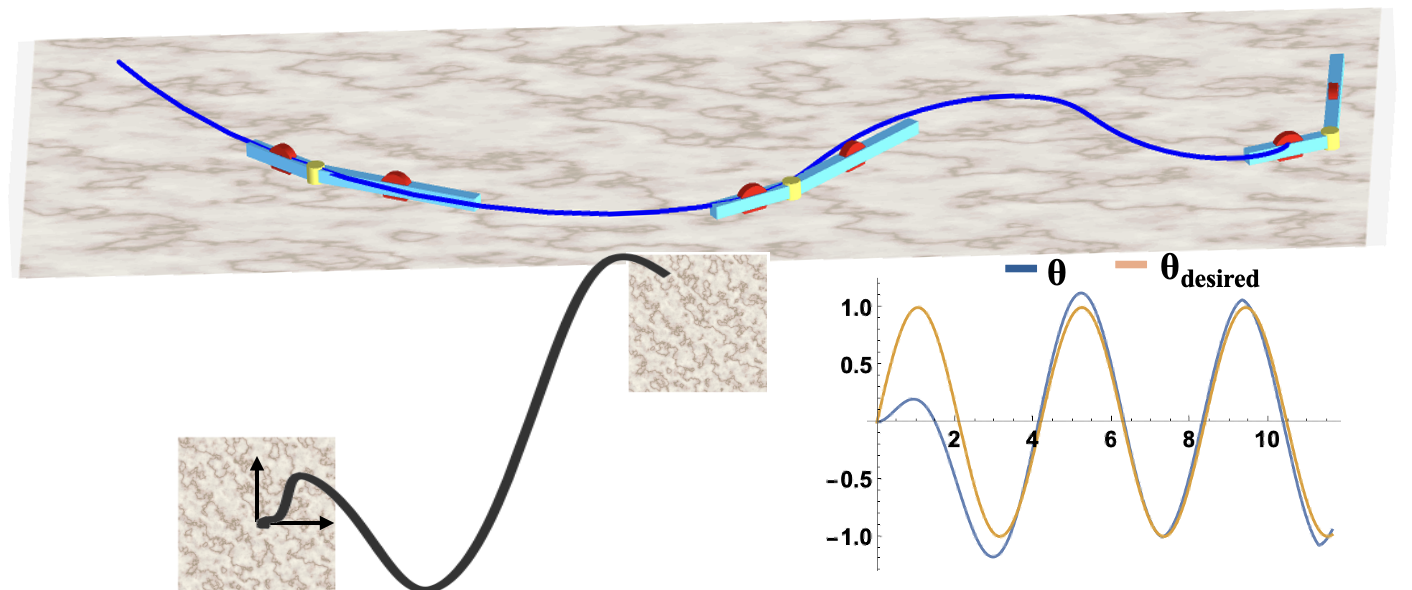}
    \caption{Stabilization of the robot to an undulatory trajectory through wavelike platform movement. The tracking of $\theta_\text{desired}(t)$ is shown at bottom right and the platform motion at bottom left.}
    \label{fig:SnakyCnt}
\end{figure}

\begin{figure}
    \centering
    \includegraphics[width=0.85\linewidth]{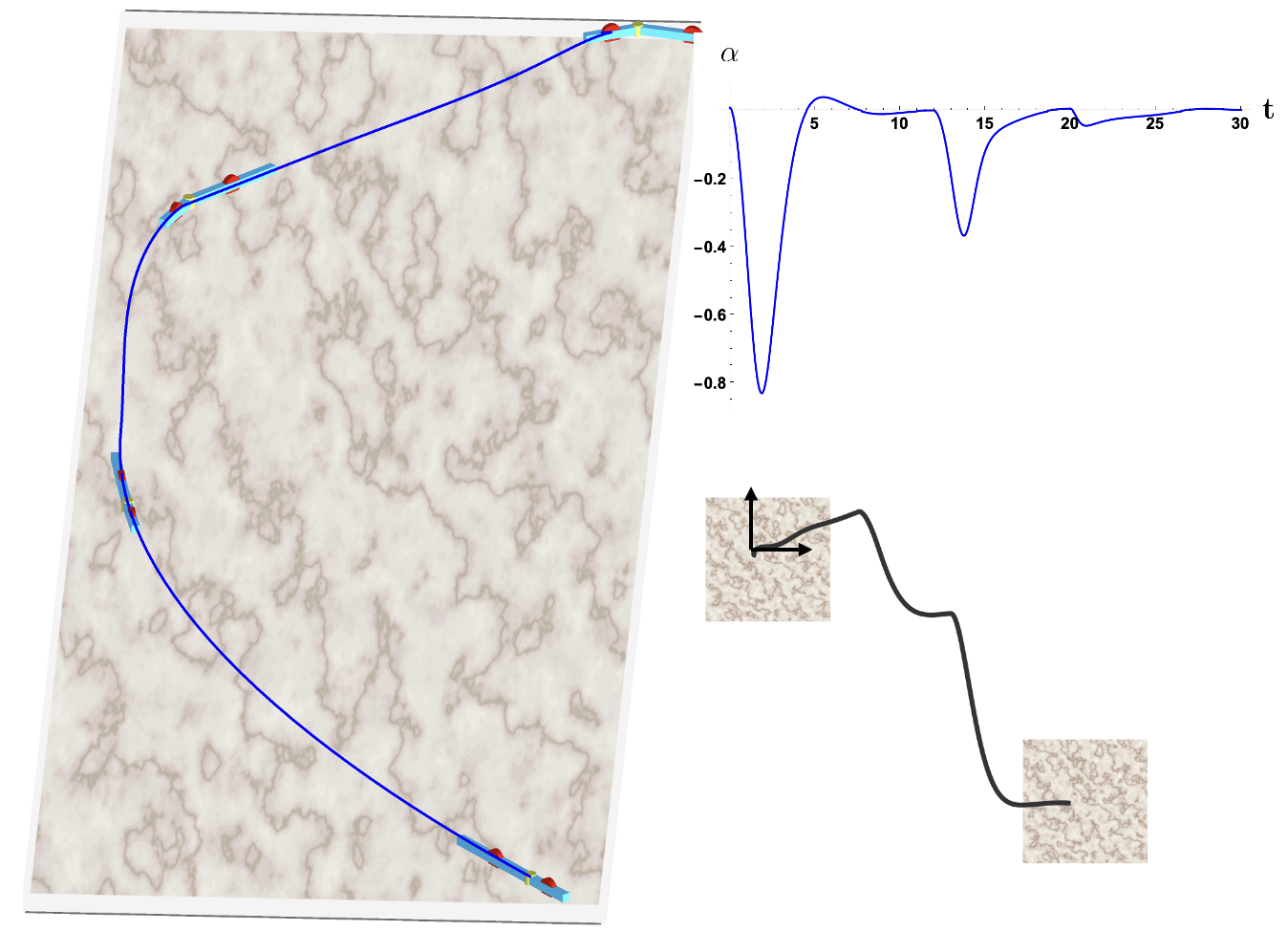}
   \caption{Sequential heading control toward multiple target points, together with joint angle history and platform motion.}
    \label{fig:MultiCnt}
\end{figure}

Figs. \ref{fig:CirbleCont} through \ref{fig:MultiCnt} demonstrate the performance of our controller.
In the first scenario, a continuously varying desired heading $\theta_\text{desired}(t)=t$ is chosen, resulting in $\dot{\theta}_\text{desired}=1$ and $\ddot{\theta}_\text{desired}=0$, so that the robot is driven to rotate with constant angular velocity and thereby traverse a circular trajectory on the platform. The middle panel of Fig.~\ref{fig:CirbleCont} shows the time evolution of $\theta(t)$ together with $\theta_\text{desired}(t)$, demonstrating accurate tracking despite the passive nature of the internal joint. The left panel depicts the resulting motion of the platform, which reflects the indirect nature of the control: as the robot approaches the aligned configuration $\alpha\approx 0$, the control effort required to maintain the desired rotational motion increases, leading to larger platform excursions. For this reason the platform trajectory has been scaled.

In the second scenario, the desired heading is specified as a smooth sinusoidal function $\theta_\text{desired}(t)$, inducing a time-varying orientation that results in an undulatory motion of the robot like that studied in Section \ref{amsSchooling.sec}.  As the wavelike locomotion pattern produces oscillation in the heading, the coupling between the platform motion and the passive joint dynamics recalls the dead fish from \cite{bhtll06}, excited to undulate --- and thus to propel itself forward --- by oscillations in its environment.
The lower right panel of Fig.~\ref{fig:SnakyCnt} illustrates the evolution of $\theta(t)$ alongside the desired trajectory $\theta_\text{desired}(t)$, demonstrating accurate tracking performance. The left-hand panel shows the corresponding platform trajectory, which reflects the oscillatory nature of the control input and the resulting periodic exchange of momentum between the robot and the platform.

In the final scenario, the heading controller is organized in a two-layer architecture for sequential point-to-point navigation on the platform. At the upper layer, the current target point determines the desired heading angle through the direction of the line joining the robot position to that target. At the lower layer, the feedback-linearizing tracking law developed above drives $\theta(t)$ toward this desired orientation; once the heading error falls below a prescribed tolerance, the controller advances to the next target and repeats the procedure. The snapshots shown in Fig.~\ref{fig:MultiCnt} depict the initial robot configuration together with the three successive target points $(-15,-15)$, $(-15,-30)$, and $(0,-60)$, illustrating the resulting piecewise reorientation and translation of the robot relative to the platform. As expected from momentum exchange, every relative displacement of the robot is accompanied by motion of the platform in the opposite direction, and each rapid transition to a new desired heading is associated with a pronounced platform excursion required to generate the reorienting impulse.

\addtolength{\textheight}{-3cm}   


\section{CONCLUSIONS AND FUTURE WORK}

The two-link robot considered above is attractive to the authors as a ``model organism'' for the study of robotic locomotion because its dynamical properties are suggestive of a biological system widely quoted in ``bio-inspired'' designs, while its equations of motion --- though rich with nonlinearity --- can be treated analytically, given the proper mathematical tools. We've shown that when multiple such robots share a common environment with its own inertial dynamics, the resulting coupling between robots can substantially impact their individual dynamics, in a manner reminiscent of the physics underlying mutualistic energy-efficient fish schooling. Within a fish school, each fish is driven externally not by its neighbors --- at least not directly --- but by the flow conditions its neighbors they create. Isolating the problem of a single two-link robot on a movable platform, we've shown that if the robot relinquishes all authority over its internal shape, control of the platform's acceleration is sufficient to afford indirect control of this internal shape, and thus of the robot's locomotion.

We conclude by noting that if a robot with a passive joint shares a free platform with at least two robots whose joints are actively controlled, then it's unquestionably possible for the active robots to dictate the platform's acceleration, and thus to control the motion of the passive robot. Recall that a front-to-back symmetric robot, when flexing its internal joint, will simply scissor back and forth relative to the platform, its center of mass remaining on the straight-line bisector of its internal joint. Two such robots atop a free platform, as long as they're not initially aligned, afford direct control over the platform's acceleration. Difficulty arises when these robots are front-to-back asymmetric, so that they begin to locomote and reorient when they perform this actuation. The controlled schooling of mixed populations of active and passive asymmetric robots is the authors' next objective.

%


%


\bibliography{references}
\bibliographystyle{ieeetran}

\end{document}